\documentclass{article}

\usepackage[preprint,nonatbib]{neurips_2023}

\usepackage[utf8]{inputenc} 
\usepackage[T1]{fontenc}    
\usepackage{hyperref}       
\usepackage{url}            
\usepackage{booktabs}       
\usepackage{amsfonts}       
\usepackage{nicefrac}       
\usepackage{microtype}      
\usepackage{xcolor}         

\usepackage{graphicx}
\usepackage{multirow}

\title{Residual Learning for Image Point Descriptors}

\author{%
Rashik Shrestha$^{1}$ \quad Ajad Chhatkuli$^{1,2}$ \quad Menelaos Kanakis$^2$ \quad Luc Van Gool$^2$ \\
$^1$NepAl Applied Mathematics and Informatics Institute for research (NAAMII) \\ \quad $^2$ETH Zürich \\
\texttt{rashik.shrestha@naamii.org.np}\\
\texttt{\{ajad.chhatkuli,menelaos.kanakis,vangool\}@vision.ee.ethz.ch}
}

\begin{document}
\maketitle
\begin{abstract}
Local image feature descriptors have had a tremendous impact on the development and application of computer vision methods. It is therefore unsurprising that significant efforts are being made for learning-based image point descriptors. However, the advantage of learned methods over handcrafted methods in real applications is subtle and more nuanced than expected. Moreover, handcrafted descriptors such as SIFT and SURF still perform better point localization in Structure-from-Motion (SfM) compared to many learned counterparts. In this paper, we propose a very simple and effective approach to learning local image descriptors by using a hand-crafted detector and descriptor. Specifically, we choose to learn only the descriptors, supported by handcrafted descriptors while discarding the point localization head. We optimize the final descriptor by leveraging the knowledge already present in the handcrafted descriptor. Such an approach of optimization allows us to discard learning knowledge already present in non-differentiable functions such as the hand-crafted descriptors and only learn the residual knowledge in the main network branch. This offers 50X convergence speed compared to the standard baseline architecture of SuperPoint while at inference the combined descriptor provides superior performance over the learned and hand-crafted descriptors. This is done with minor increase in the computations over the baseline learned descriptor. Our approach has potential applications in ensemble learning and learning with non-differentiable functions. We perform experiments in matching, camera localization and Structure-from-Motion in order to showcase the advantages of our approach.
\end{abstract}
\section{Introduction}
\label{sec:intro}
The impact of feature point localization and description methods~\cite{lowe2004distinctive,bay2006surf,rublee2011orb,alcantarilla2012kaze,calonder2010brief,tola2009daisy,detone2018superpoint} in computer vision applications cannot be understated. Key computer vision applications such as Structure-from-Motion (SfM)~\cite{schonberger2016structure,snavely2008modeling,wu2013towards} and sparse Simultaneous Localization and Mapping (SLAM)~\cite{mur2015orb,davison2007monoslam}, hinge on the remarkable accuracy of local feature descriptor methods. SfM and SLAM in turn have facilitated successful industrial and scientific applications~\cite{giubilato2018experimental,ozden2010multibody}. As an example, SfM has played no small part in the optimization of Neural Radiance Field (NERF) models by offering accurate camera poses~\cite{mildenhall2021nerf} and sparse 3D initialization~\cite{guo2022neural}.

State-of-the-art local feature descriptors are either handcrafted or learned. Handcrafted feature descriptors often use spatial gradients~\cite{mikolajczyk2001indexing,lowe2004distinctive, bay2006surf,rublee2011orb,harris1988combined,rosten2006machine} to localize interest points on images. These methods provide repeatable interest points despite view point changes and changes in scale with remarkable accuracy. The interest point detection is then followed by the descriptor computation, which often uses histogram features~\cite{lowe2004distinctive,tola2009daisy,rublee2011orb} for robustness. The descriptor computation is often performed through non-differentiable functions~\cite{lowe2004distinctive,rublee2011orb} that are not trainable in classical deep learning. The importance of such will be apparent later on.

Learned methods for local image features have gained traction through self-supervised learning~\cite{doersch2017multi} and specifically contrastive and metric learning~\cite{sun2014deep,chen2020simple} on image augmentations. A method that stands out with self-supervised training on a large image set~\cite{lin2014microsoft} is SuperPoint~\cite{detone2018superpoint}. A key advantage of the method is regarding the ease of training in terms of hardware requirement and the simple approach. Latter works have proposed improvements via score prediction~\cite{revaud2019r2d2,tang2019neural}, outlier rejection~\cite{tang2019neural} and larger transformer networks~\cite{Wang_2022_ACCV}. Furthermore, self-supervised learning on large image sets alleviates domain shift problems encountered by earlier learned methods~\cite{schonberger2017comparative}. Metric learning~\cite{sun2014deep} on the other hand, can address some of the invariance vs.\ description tradeoff. Despite these advances, a severe limitation of learned methods is in fact on the sub-pixel point localization. A challenging benchmark~\cite{jin2021image} highlights the low `resolution' of point clouds in the SfM reconstructions. \cite{tyszkiewicz2020disk} uses a more complex training approach using reinforcement learning in order to solve the problem of point localization resolution as well as providing better descriptors. However, the method requires significant computation in training and inference, requiring specialized GPUs for training. Owing to the accuracy provided by hand-crafted keypoints, earlier works~\cite{choy2016universal,mishchuk2017working} learn only the descriptors. Specifically \cite{mishchuk2017working} directly uses SIFT~\cite{lowe2004distinctive} in order to compute the interest points. The current state-of-the-art paradigm of local image point description is however, to learn descriptors at interest points identified by a point detection branch trained jointly with the descriptor network.

\setlength{\belowcaptionskip}{-13pt}

\begin{figure*}[t]
    \centering
    \includegraphics[width=1\textwidth]{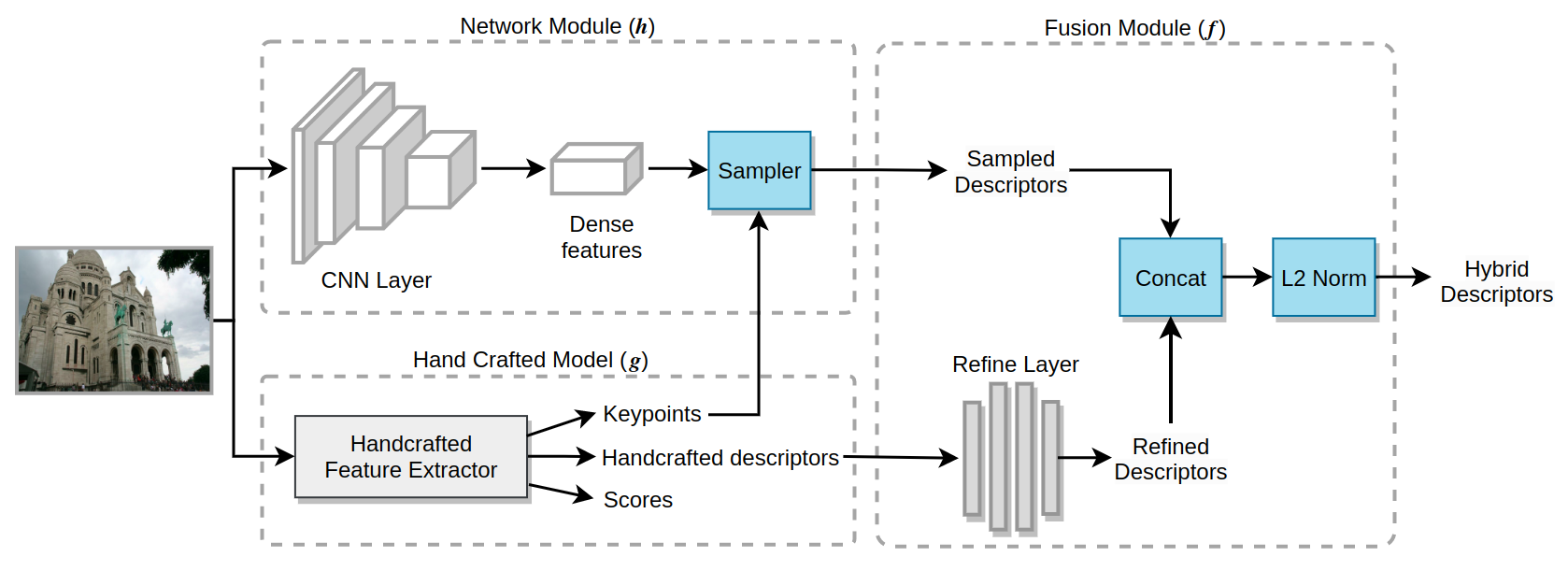}
    \caption{Block Diagram of our approach, consisting three main blocks: Network Module $h$, Hand-crafted Model $g$ and Fusion Module $f$. Despite the hand-crafted part being non-differentiable, our method can make use of the function for learning residual knowledge on the network modules.}
    \label{fig:block_diagram}
\end{figure*}

\setlength{\belowcaptionskip}{0pt}

Two questions naturally arise with these observations. Can traditional methods~\cite{lowe2004distinctive, bay2006surf} provide further supervision to the task of self-supervised learning of description networks such as the Superpoint~\cite{detone2018superpoint}, while exploiting the high accuracy of hand-crafted point localization? In other words, can we \emph{improve} the hand-crafted descriptors in conjunction with a standard learning approach for image point descriptors? More importantly, given the low-compute advantage of the handcrafted methods, can these two different approaches be used collaboratively to construct a more powerful description of interest points? In this work, we address the above questions by constructing a simple albeit non-conventional learning architecture. We aim to maximize the use of knowledge encoded by a handcrafted method and learn the residual knowledge in the Deep Neural Nets (DNNs). Here we use `residual knowledge' in a broader sense for describing knowledge learned on top of existing ones, and not just for a summation operation. Similar to \cite{dusmanu2021cross}, we project the handcrafted feature using a learnable multi-linear perceptron (MLP) followed by concatenation with the DNN feature. The simple architecture forces the DNN to only learn `residual information' on the network.
Thus the whole method makes efficient use of arbitrary but useful non-differential functions for the task. We highlight our approach in Figure~\ref{fig:block_diagram}. We perform extensive evaluations with ablations that showcase superior performance of the proposed method in various metrics including localization and reconstruction.

\section{Related Work}
\label{sec:relwork}

We discuss briefly the relevant works on learning feature extractors. Specifically, we are interested in learned local feature extractors. We divide them into those which do not fully learn keypoints and descriptors and those that learn both the keypoints and descriptors.

\paragraph{Learned methods with keypoints or descriptors.}
Hand-crafted features provide the advantage of speed and accuracy of point localization. Thus earlier methods~\cite{choy2016universal,mishchuk2017working} instead learn only the descriptors. Instead of learning the descriptors, it is also possible to directly learn matching between two images as done in \cite{sun2021loftr,truong2022probabilistic}.

\paragraph{Learned methods with local features and keypoints.}
Fully learned local features provide an attractive incentive to the research community in the hope of maximizing the data priors in methods. TILDE~\cite{verdie2015tilde} and LIFT~\cite{yi2016lift} are seminal works in that regard which use ground-truth matches in order to learn descriptors. Next, several methods proposed self-supervised or unsupervised formulation for training descriptors~\cite{choy2016universal,lenc2016learning,zhang2018learning,detone2018superpoint}. A key challenge with any learned descriptor is the domain gap at inference that naturally arises when testing on images of the user's choice. This has been partly solved by using augmentations and learning in a very large image set, together with metric learning~\cite{detone2018superpoint}. Nonetheless, the question of perspective changes from different viewpoints still remain very recently tackled by \cite{Muhle2023LearningCU}. Several recent works have provided architectural improvement~\cite{lee2022self} and novel training loss~\cite{tyszkiewicz2020disk}. However, in practice handcrafted approaches such as SIFT~\cite{lowe2004distinctive} and standard learned approach of \cite{detone2018superpoint} remain highly attractive methods in the community.

\section{Method}
\label{sec:method}

In order to formalize our approach, we consider two functions $g(\mathbf{I})\to \{(x,\ y_1)\}, y_1 \in \mathbb{R}^{d_1}$ -- the non-differentiable model which is the handcrafted feature descriptor function and the learned descriptor function $h(\mathbf{I}, \{x\}) \to \{y_2\} \in \mathbb{R}^{m\times d_2}$. The hand-crafted model $g$ takes in an image and outputs $m$ keypoint locations $\{x\}$, along with a descriptor of dimension $d_1$ for each point. The learned feature descriptor function $h$ uses the image and the point locations $\{x\}$ to output descriptors $\{y\}$, each of dimension $d_2$. Here, $(x,y_2)$ is a tuplet of point location and corresponding descriptor respectively. Additionally, a feature fusion module $f(\{(y_1, y_2)\})\to \{y\}, y \in\mathbb{R}^{d}$ takes in the two corresponding feature set outputs of $f$ and $g$ in order to output $m$ descriptors of dimension $d$. In our case, only $h$ and $f$ are trainable functions, implemented as neural networks. Our goal is to learn the functions $h$ and $f$ in order to produce descriptors $y$ which has better performance compared to $y_1$ or the descriptor obtained by training $f$ independently of $g$.

The above formalization helps us to understand the problem as that learning $h,\ f$ conditioned on $g$, rather than only feature fusion. In semi-supervised learning~\cite{lee2013pseudo} the use of two different functions $g$ and $h$, where one is an expert is used to train the network $h$ efficiently through the so-called self-learning. Two important differences exist in our case, the expert $g$ is computationally more efficient than $h$ but is not sufficiently good. Furthermore, we have an effective approach of metric learning in order to supervise $h$ and thus $f$ similar to previous works~\cite{detone2018superpoint,revaud2019r2d2,tang2019neural,maurer2022zippypoint}. 

\subsection{Framework Design}
Our framework is divided into three parts: the non-differentiable function $g$, the network module $h$ and the fusion/projection module $f$. We illustrate their constructions and interactions through Figure~\ref{fig:block_diagram}. Below we detail each of the block and their design choices.

\paragraph{Handcrafted method $g$.}
We mainly consider the two most popular full precision handcrafted descriptor methods: SIFT~\cite{lowe2004distinctive,arandjelovic2012three} and SURF~\cite{bay2006surf}. Both use gradient information for detection of interest keypoints. The main differences between the two, from the design perspective, are the speed and feature size they return. SIFT is computationally more complex and uses a descriptor length of 128 integers. SURF on the other hand, is faster and uses float vectors which can be of lengths 32 or 64. Although these two methods may be combined with yet another fusion module, we choose to use only one of them in order to focus on analysis over the performance.

\paragraph{Neural network $h$.}
We use a standard architecture~\cite{detone2018superpoint} for learning the dense descriptors in an image. The network $h$ takes in an image $\mathbf{I}$ to compute dense descriptors. Before they are used in the loss function, these dense descriptors are sampled at locations $\{x\}$, provided by the handcrafted method $g$. \cite{detone2018superpoint} uses VGG style encoder layer to reduce the dimensionality of input image and descriptor decoder head to generate final descriptors.

\paragraph{Fusion module $f$.} Feature fusion is often used in prior works in order to combine complementary features~\cite{hou2017dualnet} or for knowledge distillation~\cite{hinton2015distilling}, to name a few. We use a small 3-layer MLP which we call the Refine layer in order to project the handcrafted descriptors obtained from $g$. We then concatenate the output of the refine layer and the network module $h$ to obtain the final descriptors. Note that alternate design choices can be used instead of concatenation, for example, addition or element-wise product. However, we opt for concatenation for simplicity and favorable results from our initial experiments. We use $\ell_2$ normalization in order to bound the descriptors after the concatenation.

\subsection{Network Architecture}

Our setup uses two learning networks, one CNN layer to extract deep features from the image $h$ and one small Refine layer $f$ to project the handcrafted features before concatenation.

\begin{figure}
    \centering
    \includegraphics[width=0.5\textwidth]{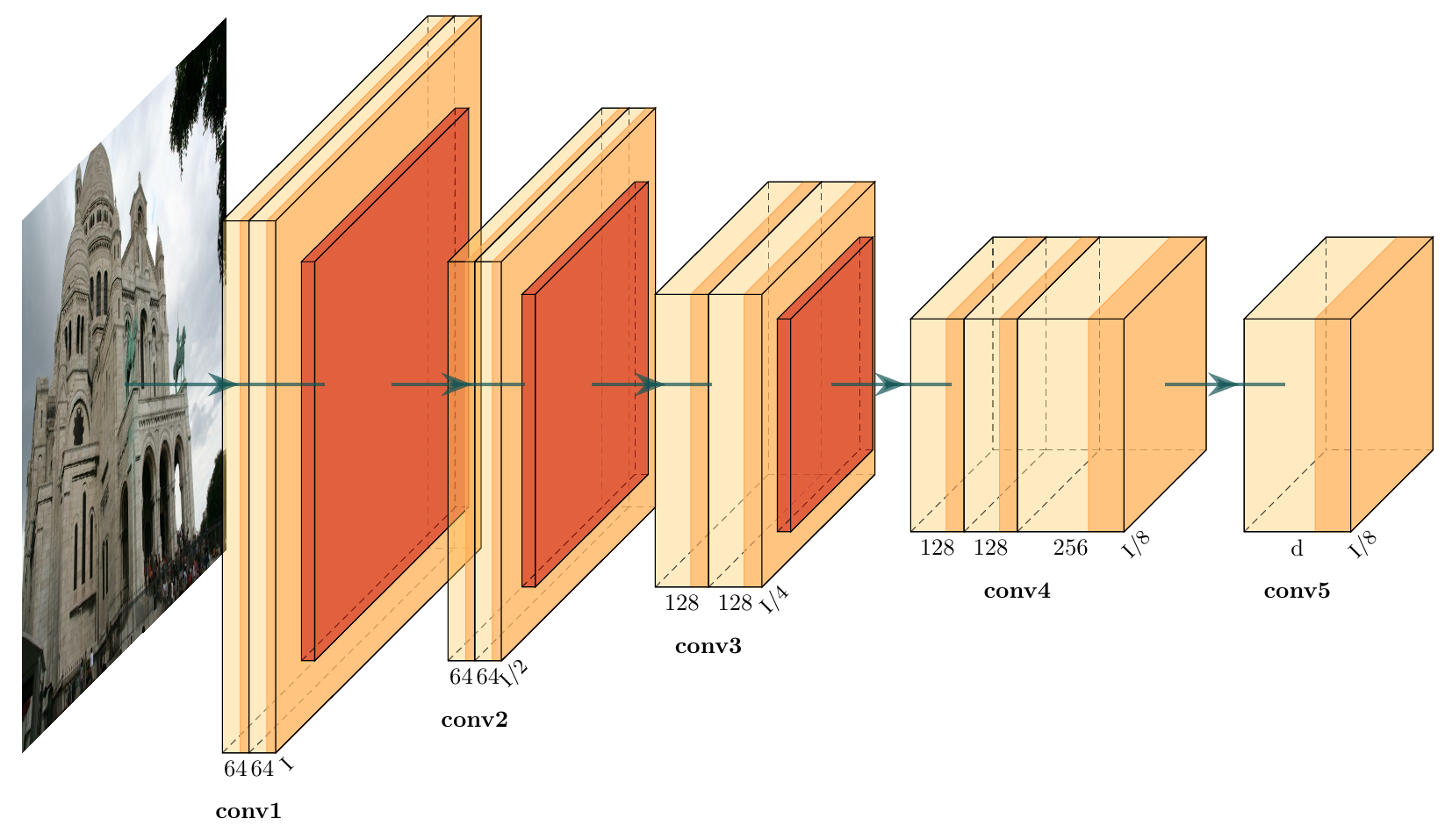}
    \caption{Trimmed SuperPoint Architecture: We use the SuperPoint~\cite{detone2018superpoint} as our baseline architecture for $h$ without the point decoder head.}
    \label{fig:trimmed_sp}
\end{figure}

For the \textbf{CNN layer}, we use a trimmed version of the standard SuperPoint architecture by removing its point decoder head. It has a VGG style encoder layer to reduce the dimensionality of input image and descriptor decoder head to generate the dense descriptors. We use RGB image for input, so for an image of size $3 \times H \times W$, the network generates features of dimension $d_s \times H_c \times W_c$ dimension, where $H_c = H/8 $ and $W_c = W/8$. Figure \ref{fig:trimmed_sp} shows the architecture in details. 

For the \textbf{Refine layer}, we use a simple MLP with two hidden layers of size $256$, making the architecture $d_h \rightarrow 256 \rightarrow 256 \rightarrow d_r$. 

In our experiments, we use $d_s = 128$ and $d_r=128$. For SIFT and extended SURF descriptors, $d_h=128$. We use simple ReLU activation after CNN and linear layers. Note that several other design choices exist for the Refine layer and the feature fusion thereafter. We do not use additional layers and keep the Refine layer small as per the results of our initial experiments.

\subsection{Self-Supervised Training}
We train all models including \cite{detone2018superpoint} in the large MS COCO~\cite{lin2014microsoft} dataset, using the self-supervised learning proposed in \cite{detone2018superpoint}. Specifically, \cite{detone2018superpoint} uses metric learning on augmentations that combine geometric and non-geometric transformations for robust learning.
The loss function is the following:
\begin{equation}
\label{eq:loss}
\mathcal{L} = \sum_{i} max(0,  y(a,p) - y(a,n) + m)
\end{equation}
Here, $y$ is the final descriptor of a method with $(a,p)$ denoting a positive for an anchor point $a$ and $(a,n)$ denoting a negative for the same anchor point.
We use the margin of $m=2$, which is the maximum possible distance for the normalized vectors.

\section{Experiments}
\label{sec:exp}
In this section, we provide the necessary implementation details, experimental setups and evaluations.

\begin{figure}
    \centering
    \includegraphics[width=\textwidth]{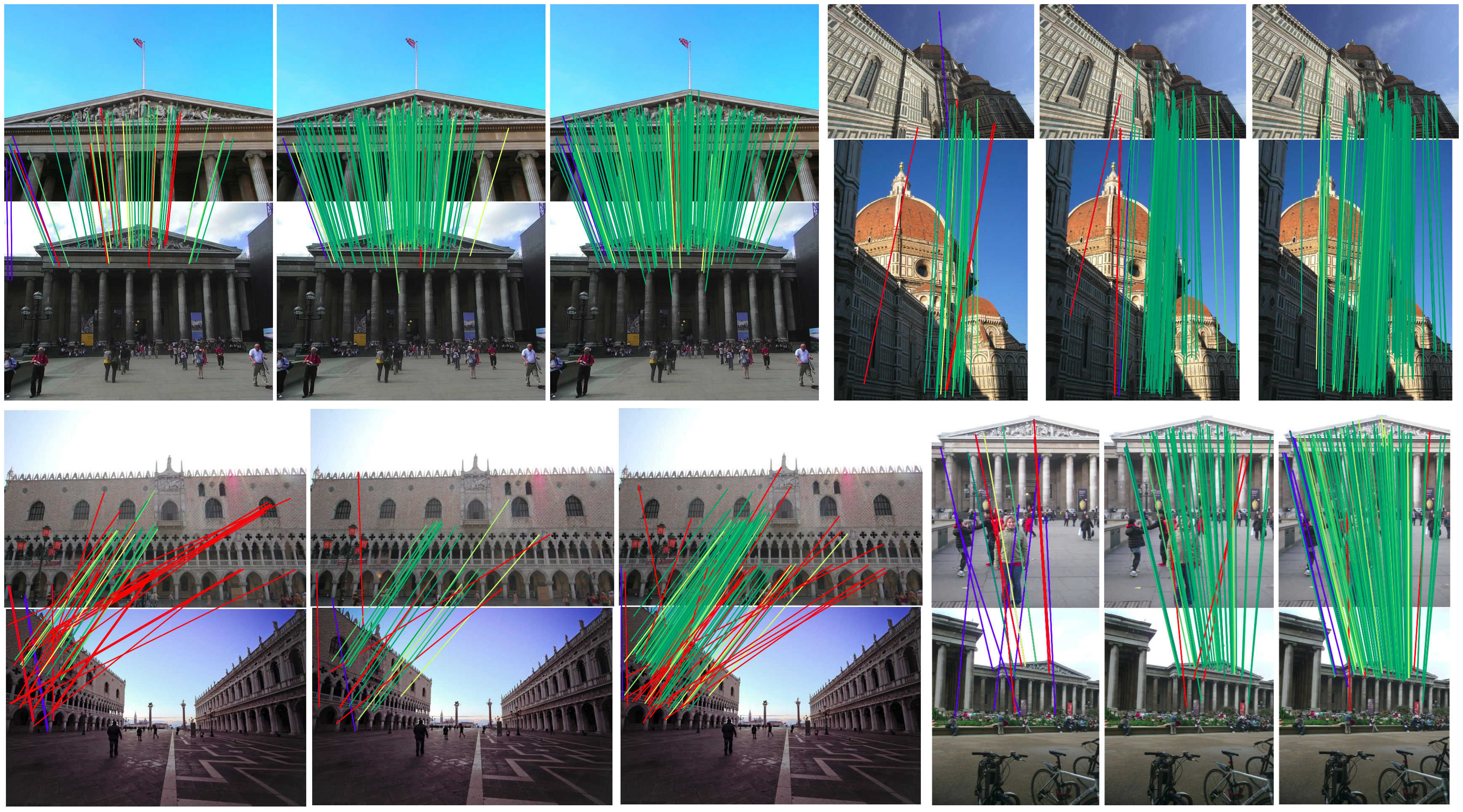}
    \caption{\textbf{Matches Visualization.} We visualize the matches produced by Superpoint (left), Upright RootSIFT (middle), and our method (right) for four stereo pairs in the test set of Phototourism dataset. Matches are coloured red to green, according to their reprojection error (high to low).}
    \label{fig:match_vis}
\end{figure}

\begin{figure}
    \centering
    \includegraphics[width=\textwidth]{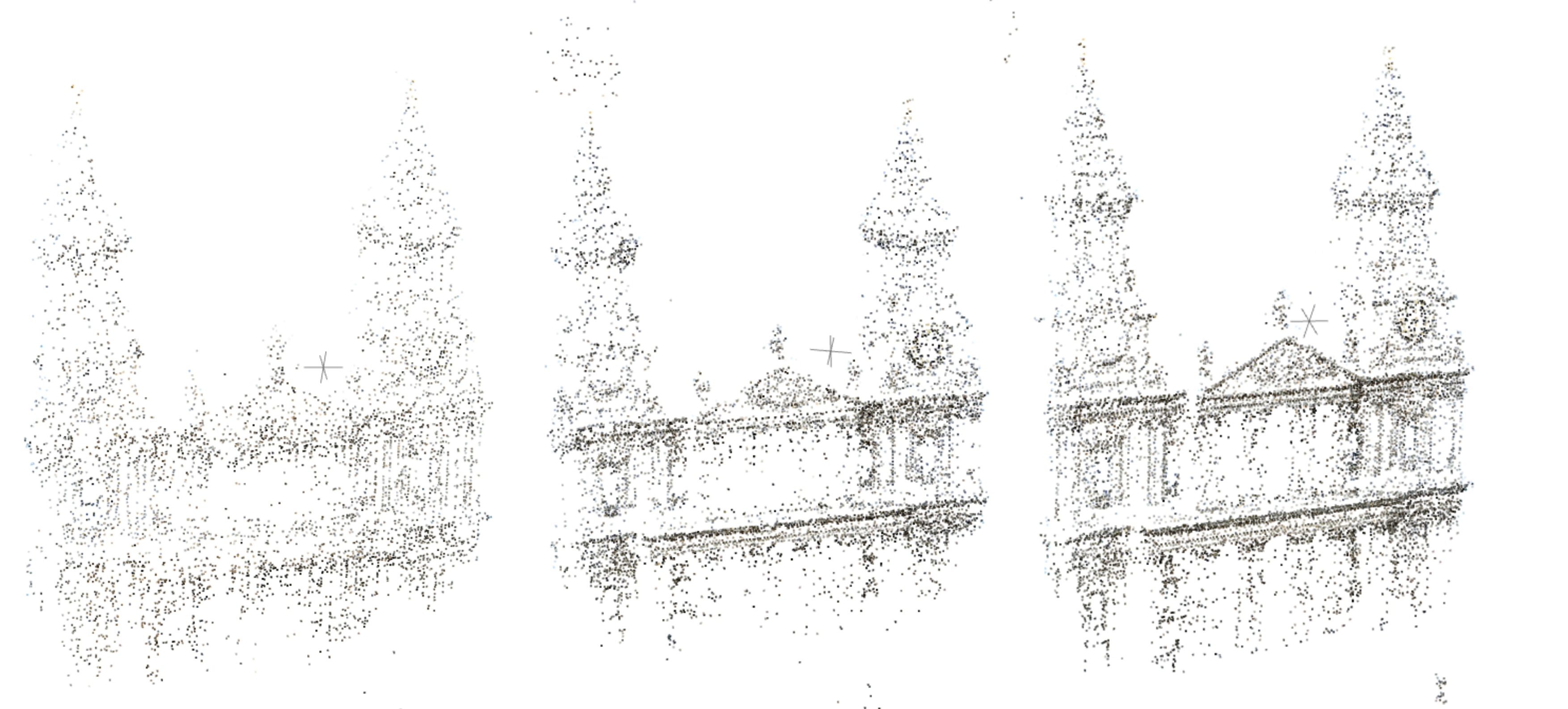}
    \caption{\textbf{Point Cloud Visualization.} We build SFM model of ``St Pauls Cathedral'' scene of Phototourism dataset using 25 images. We visualize point clouds produced by Superpoint, RootSIFT Upright, and our method (from left to right), having 7k, 8k and 11k landmarks respectively.}
    \label{fig:pointcloud}
\end{figure}

\subsection{Implementation Details}

\paragraph{Train.} Following the common training paradigm of \cite{detone2018superpoint}, we use MS-COCO 2017 training dataset split having $118k$ images to train our model. Each image is warped using a random homography transformation. Known homography allows us to generate a set of corresponding keypoints in the original image and their respective warped counterparts, which serves as baseline matches to train our model. Moreover, we use standard non-geometric data augmentation techniques such as addition of random Gaussian noise, random changes to brightness, contrast, saturation and hue to improve the network’s robustness.

We implement the model using PyTorch framework. We use a batch size of $4$ and ADAM optimizer. The learning rate was set to $10^{-3}$ throughout the training. We optimize the model for just $2$ epochs, which is significantly less than alternative learned methods, such as 100 for SuperPoint, since we found that longer training did not significant improvement the model's performance.

\paragraph{Test.} We test our model on HPatches dataset~\cite{balntas2017hpatches} for Homography estimation task, AachenV1.1 Day night dataset \cite{sattler2018benchmarking, sattler2012image} for Pose estimation task, and on the Image matching challenge for stereo and multi-view tasks. A more detail explanations can be found in their corresponding sections. 

\subsection{HPatches}
\label{sec:hpatches}
\begin{table*}[ht]
\centering

\begin{tabular}{l|cccc|cccc}
\toprule

\multirow{2}{*}{Models} & \multicolumn{4}{c|}{240x320 - 300 points } & \multicolumn{4}{c}{480x640 - 1000 points} \\
& Cor-1$\uparrow$ & Cor-3$\uparrow$ & Cor-5$\uparrow$ & MS$\uparrow$ & Cor-1$\uparrow$ & Cor-3$\uparrow$ & Cor-5$\uparrow$ & MS$\uparrow$ \\

\midrule
    
ORB             & 0.112 & 0.369 & 0.474 & 0.206 & 0.217 & 0.528 & 0.621 & 0.200 \\
SIFT            & 0.583 & 0.841 & 0.884 & 0.268 & \textbf{0.590} & \textbf{0.867} & 0.914 & 0.289 \\
SURF            & 0.397 & 0.702 & 0.762 & 0.255 & 0.421 & 0.745 & 0.812 & 0.230\\
SuperPoint      & 0.491 & 0.833 & 0.893 & 0.318 & 0.509 & 0.834 & 0.900 & 0.281\\
Ours (SIFT)     & \textbf{0.498} & \textbf{0.868} & \textbf{0.905} & 0.400 & 0.547 & 0.859 & \textbf{0.933} & \textbf{0.407} \\
Ours (SURF)     & 0.492 & 0.821 & 0.901 & \textbf{0.401} & 0.531 & 0.821 & 0.917 & 0.401 \\

\bottomrule
\end{tabular}

\caption{Descriptor evaluation on HPatches dataset. The table shows comparison of Matching Scores and Homography estimation accuracy with 3 different pixel distance thresholds (1,3 and 5). We highlight the best method in \textbf{bold}.}
\label{tab:hpatches_metrics}
\end{table*}

Hpatches is comprised of 116 scenes, each with 6 images. All images in a scene are related by a known homographic transformation. 57 scenes have illimunation variations and 59 has viewpoint variations. We report descriptor evaluation metrics, namely Matching Score (MS) and Homography Accuracy (Cor) with thresholds of 1, 3 and 5 pixels. The first image of each scene is used as a reference image to match with the other 5. Hence we have 116*5=580 match pairs.

We perform two sets of experiments with low and high resolution images. For low-resolution images of size 240x320, number of keypoints is limited to 300. For high-resolution images of size 480x640, number of keypoints is limited to 100. Our approach gave better results than SuperPoint and other handcrafted method. We find that extending both SIFT or SURF with our method, consistently improves performance to the handcrafted alternative as well as when using just the SuperPoint baseline. This is thanks to the residual learning of additional knowledge in the deep network branch not captured by the hand-crafted descriptors.

\subsection{Aachen Day Night}
\label{sec:aachen_day_night}
\begin{table*}[ht]
\centering

\begin{tabular}{l|ccc|ccc}
\toprule

\multirow{4}{*}{Models} & \multicolumn{6}{c}{Localization Accuracy $\uparrow$} \\
& \multicolumn{3}{c}{Day} & \multicolumn{3}{c}{Night} \\
& .25 & .5  & 5    & .25 & .5  & 5   \\
& 2  & 5   & 10    & 2  & 5  & 10 \\

\midrule

SIFT                    & 82.3  & 91.6  & 97.0  &  45.0 & 58.6  & 72.8  \\
SURF                    & 83.4  & 91.7  & 96.6  &  41.4 & 56.0  & 70.7  \\
SuperPoint              & 86.8  & \textit{93.8}  & \textbf{97.9}  & \textit{62.3}  & \textbf{81.7}  & \textit{94.8}  \\

\midrule

Ours \footnotesize (SIFT)     & \textbf{87.0}  & \textbf{94.8}  & \textbf{97.9}  & \textit{62.3}  & 80.6  & \textbf{95.3}  \\
Ours \footnotesize (SURF)     & \textit{86.9}  & 93.3  & \textit{97.7}  & \textbf{63.4}  & \textit{81.2}  & \textit{94.8} \\

\bottomrule
\end{tabular}

\caption{Visual Localization accuracy on Aachen v1.1 dataset with different threshold values for day and night time images. We highlight the best method in \textbf{bold} and \textit{italicize} the second-best.}
\label{tab:hloc_metrics}
\end{table*}

To further investgiate the generalization capabilities of our method, we evaluate the efficacy of our model on AachenV1.1 \cite{sattler2012image}.
AachenV1.1 is comprised of challenging real-life images captured during different times of the day, namely, day and night.
Specifically, the dataset has 824 daytime and 191 nighttime query images. 
We report the localization accuracy with the threshold values of ($0.25m$ ,$2^{\circ}$), ($0.5m$ ,$5^{\circ}$) and ($5m$ ,$10^{\circ}$). 
We utilize the HLoc~\cite{sarlin2019coarse} framework for localization, and for each query image, we retrieve the 30 closest images based on global descriptors extracted using NetVLAD~\cite{arandjelovic2016netvlad}. These images act as localization candidates for the query image to localize within the 3D map.

We compare our model with standard handcrafted methods, namely SIFT and SURF, while also evaluating the baseline learned method SuperPoint. For SuperPoint, we used provided pre-trained weights. We test our model with SIFT and SURF features.

We lower the detection thresholds in all our experiments to get enough points. For SIFT, contrast threshold is set to $0.005$. For SURF, hessian threshold is set to $30$. For SuperPoint, keypoint detection threshold is set $10^{-4}$. For all, we take only top $4096$ points, filtered on the basis of keypoint scores.

Our method with SIFT worked better than others for Day time images. We can see that handcrafted methods SIFT and SURF does not show good performance for night time images. But a small boost from our model has significantly improved their performance, better or comparable to that of Superpoint. Our model has about 20\% of boost in improvement as compared to their hand crafted counterparts for night time images. The difference is more significant when the threshold value is low.

Table \ref{tab:hloc_metrics} shows the results of our experiment. Our method with SIFT features worked consistently better in all the day time metrics.

\subsection{Image Matching Challenge (IMC)}
\label{sec:image_matching_challenge}
\begin{table*}[ht]
\centering

\resizebox{\columnwidth}{!}{%
\begin{tabular}{ll|cccc|ccccc}

\toprule

& & \multicolumn{4}{c|}{Stereo Task} & \multicolumn{5}{c}{Multiview Task} \\
& Models & NM & NI & mAA(5\textdegree) & mAA(10\textdegree) & NM & NL & TL & mAA(5\textdegree) & mAA(10\textdegree) \\

\midrule

\multirow{7}{*}{\rotatebox[origin=c]{90}{\small 2k keypoints}} 
& RootSIFT & 163.1 & 86.0 & 0.2161 & 0.3112 & 164.1 & 1179.5 & 3.78 & 0.3712 & 0.4699 \\
& RootSIFT Upright & 197.4 & 114.6 & \textit{0.2619} & \textit{0.3658} & 201.0 & 1400.6 & 4.06 & \textit{0.4333} & \textit{0.5413} \\
& SURF & 145.9 & 56.3 & 0.1277 & 0.2020  & 147.8 & 900.2 & 3.38 & 0.2476 & 0.3269 \\
& SuperPoint & \textit{211.6} & 103.2 & 0.2045 & 0.3015 & \textit{215.7} & 1392.7 & 4.24 & 0.3982 & 0.5117 \\
& Ours \small(RootSIFT) & 185.5 & \textit{116.3} & 0.2426 & 0.3424 & 187.4 & \textit{1454.2} & \textit{4.10} & 0.4142 & 0.5235 \\
& Ours \small(RootSIFT Up) & \textbf{216.7} & \textbf{134.7} & \textbf{0.2766} & \textbf{0.3852} & \textbf{219.6} & \textbf{1582.8} & \textbf{4.14} & \textbf{0.4575} & \textbf{0.5713} \\
& Ours \small(SURF) & 160.2 & 103.9 & 0.1907 & 0.3125 & 159.0 & 1248.2 & 4.01 & 0.3974 & 0.4517 \\

\midrule

\multirow{7}{*}{\rotatebox[origin=c]{90}{\small 8k keypoints}}
& RootSIFT & 613.3 & 327.8 & 0.3667 & 0.4837 & 624.9 & 4526.9 & 4.12 & 0.5699 & 0.6731 \\
& RootSIFT Upright & 528.0 & 360.5 & 0.3940 & 0.5120 & 544.4 & 4417.6 & 4.36 & 0.5755 & 0.6760 \\
& SURF & 370.2 & 118.4 & 0.1594 & 0.2402 & 375.6 & 2439.7 & 3.38 & 0.3001 & 0.4099 \\
& SuperPoint & 599.5 & 273.6 & 0.2667 & 0.3546 & 406.8 & 3556.8 & 3.87 & 0.3900 & 0.5022 \\
& Ours \small(RootSIFT) & \textit{607.8} & \textit{397.7} & \textit{0.4020} & \textit{0.5265 } & \textit{626.1} & \textit{5130.1} &\textit{ 4.42} & \textbf{0.6084} & \textbf{0.7128} \\
& Ours \small(RootSIFT Up)& \textbf{884.3} & \textbf{561.3 }& \textbf{0.4279} & \textbf{0.5524} &  \textbf{910.0} & \textbf{6287.6} & \textbf{4.50} & \textit{0.6062} & \textit{0.6994} \\
& Ours \small(SURF) & 449.1 & 210.7 & 0.2914 & 0.4215 & 514.7 & 4025.8 & 4.19 & 0.5567 & 0.5941 \\ 

\bottomrule

\end{tabular}}

\caption{\textbf{Image Matching Challenge results.} We report mean Average Accuracy at the threshold of 5\textdegree and 10\textdegree. Also, we report Number of Matches (NM), which is fed to RANSAC for stereo task and COLMAP for Multiview task. For stereo task, we report number of inlier matches (NI) after passing through RANSAC. For multiview task, we report number of landmarks (NL) and track length (TL). We highlight the best method in \textbf{bold} and \textit{italicize} the second-best, for each keypoint category (2k and 8k). }
\label{tab:imc_metrics}

\end{table*}

We evaluate our method on a benchmark provided by Image Matching Challenge~\cite{jin2021image}. It assesses the effectiveness of local features in two tasks: stereo matching and multi-view reconstruction. 

In stereo matching, local features of an image pair is matched and fed to RANSAC~\cite{fischler1981random}, for calculating their relative pose. In multi-view, COLMAP~\cite{schonberger2016structure} is used to construct SFM model using subset of 5, 10, and 25 images. Both tasks measure the performance in terms of the quality of the estimated poses, by using mean average accuracy (mAA) at a 5\textdegree and 10\textdegree error threshold. 

\paragraph{Hypermarameter Tuning.} The accuracy is very sensitive to multiple tunable hyperparameters of the pipeline. We tune the ratio threshold of ratio test and inlier threshold of DEGENSAC\cite{DEGENSAC} to get the best performance for each method.  We tune the hyperparameters using the validation set of Phototourism dataset. It has three scenes with a total of 274 images. 

\paragraph{Fine Tuning Model.} We fine-tune our model on the training set of the Phototourism dataset. It has images from 10 different scenes. We randomly sample 120k image pairs and fine-tune our model in a single epoch. 

\paragraph{Test.} We use the phototourism test set for testing. It has images from 9 scenes. For stereo matching, we calculate feature point matches between each possible image pair for every scene. For multi-view, we construct 10 SFM models with 5 images, 5 SFM models with 10 images and 3 SFM models with 25 images, for each scene. We used ratio threshold of 0.94 and inlier threshold of 0.5 in our models.

Our Model using RootSIFT Upright descriptor performed better than SuperPoint and other handcrafted methods. Figure \ref{fig:match_vis} shows the comparison of matches produced by three different methods. We see, our method produces better matches in comparison to the other two. In the case of multi-view reconstruction, point cloud given by Superpoint is more dispersed than the one produced by SIFT detector, and hence doesn't produce a good 3D map.

Moreover, our method can be thought of a way to add more performance to handcrafted descriptors. When applied to an already better RootSIFT Upright, the results are better as well. Whereas, when applied to slightly less performing SURF, our model produced slight worse results.

\subsection{Ablation Study}

\begin{figure}
    \centering
    \includegraphics[width=0.7\textwidth]{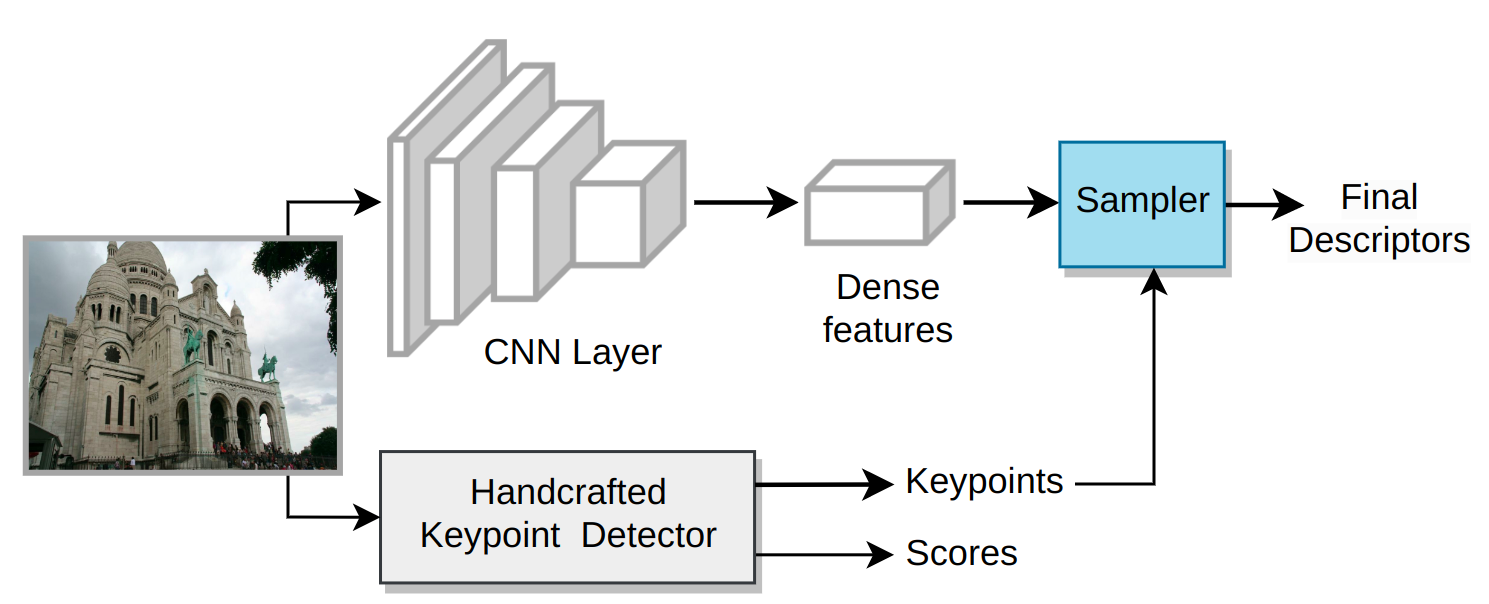}
    \caption{\textbf{Ablation Pipeline.} Fusion Module $f$ is removed}
    \label{fig:ablation_block}
\end{figure}

\begin{table*}[ht]
\centering

\begin{tabular}{l|ccc|ccc}
\toprule
\multirow{4}{*}{Models} & \multicolumn{6}{c}{Localization Accuracy $\uparrow$} \\
& \multicolumn{3}{c}{Day} & \multicolumn{3}{c}{Night} \\
& .25 & .5 & 5 & .25 & .5 & 5 \\
& 2\textdegree & 5\textdegree & 10\textdegree & 2\textdegree & 5\textdegree & 10\textdegree \\

\midrule
Ours \footnotesize (SIFT)* & 83.7 & 91.7 & 97.1	& 56.0 & 75.9 & 90.1 \\
Ours \footnotesize (SURF)* & 83.6 & 92.2 & 97.2	& 58.1 & 76.4 & 90.6 \\

\midrule

Small \footnotesize (SIFT)  & 86.8 & 93.9 & 98.4 & 60.7 & 78.5 & 91.6 \\
Small \footnotesize (SIFT)* & 84.0 & 92.0 & 96.6 & 55.5 & 71.2 & 86.4 \\

\bottomrule
\end{tabular}

\caption{\textbf{Ablation Study -} Here, * represents the model without fusion module. "Small" is the model half the size of our original model, made by removing random CNN layers.  }
\label{tab:ablation_study}

\end{table*}

The ablation study in Table~\ref{tab:ablation_study} shows that descriptors given by hand crafted model $g$ is indeed required for the model to learn quickly and generate better descriptors. We completely omit the handcrafted descriptors from the pipeline. We sample the dense features produced by network module $h$ using handcrafted keypoints, and obtain final descriptors, as illustrated in Figure~\ref{fig:ablation_block}. We omit the fusion module $f$ as well since there is only a single descriptor. 

Note that the descriptor size is kept constant as before i.e complete $256$ dimension of the descriptor is extracted from CNN Layer. Previously, first half of descriptors ($128$ dim) was obtained from CNN network, whereas last half ($128$ dim) was obtained from handcrafted method. 

We show that, concatenated descriptors perform better than CNN-only descriptors by a good margin. 

In other ablation, we study the effect of using lighter model for CNN layer. We use a smaller model with half the size than our previous one. The lighter model perform comparable to of the pretained SuperPoint model for day time images with only slight less accuracy for night time images.

\begin{table*}[ht]
\centering

\resizebox{\columnwidth}{!}{%
\begin{tabular}{ll|cccc|ccccc}
\toprule

& & \multicolumn{4}{c|}{Stereo Task} & \multicolumn{5}{c}{Multiview Task} \\
& Method & NM & NI & mAA(5\textdegree) & mAA(10\textdegree) & NM & NL & TL & mAA(5\textdegree) & mAA(10\textdegree) \\

\midrule

2k & Ours \small(SIFT) & 148.6 & 46.3 & 0.0770 & 0.1295 & 156.2 & 1578.5 & 4.21 & 0.3214 & 0.4126 \\
8k & Ours \small(SIFT) & 697.2 & 218.7 & 0.1584 & 0.2492 & 711.5 & 5405.7 & 3.98 & 0.3336 & 0.4640 \\

\bottomrule
\end{tabular}}

\caption{\textbf{Ablation study in Image Matching Challenge.}}
\label{tab:imc_ablation}

\end{table*}

We do the same ablation for image matching challenge as well, both for 2k and 8k category as shown in Table~\ref{tab:imc_ablation}. We found a huge drop in performance when excluding the handcrafted descriptors. Also, the convergence rate is slower as the model needs to learn the entire descriptor from scratch. This ablation performs as the SuperPoint model but instead of learning points on its own, it uses SIFT points. This gives the advantage for the network to learn faster, but it is still slow than training by incorporating the handcrafted descriptors. 

\subsection{Limitations}

Though our model is small in size, it still adds some computational overhead to the handcrafted descriptors. The increase in computation over the deep neural network SuperPoint is about 10\%. Application fields such as robotics may require realtime feature extraction and matching using low-powered devices. Our solution might not be ideal for such a scenario without network quantization. Although the approach we propose is general enough to learn a DNN in conjunction with multiple hand-crafted descriptors, or for that matter, a mixture of hand-crafted and pre-trained model, a stringent requirement is that they must share the same keypoint extractor. 
Another drawback of our approach is that the final concatenation increases the dimension of the descriptors which results in slightly longer matching time.
\section{Conclusions}
\label{sec:conclusion}
 
In this paper, we studied the problem of self-supervised learning of descriptors conditioned on hand-crafted descriptors. We opted a simple training strategy and architecture owing the the baseline method's success and showcased how the `knowledge' encoded in hand-crafted descriptors can be augmented by learning through a deep network. We call the knowledge added by the deep neural network as residual knowledge as it naturally avoids learning the same function as the hand-crafted descriptor. We obtained a significant improvement in performance on several challenging datasets and evaluation tasks compared to our baselines. Such an approach may be used in future in order to incorporate useful non-differentiable functions in the `residual learning' paradigm for different tasks.

{
\small
\bibliographystyle{unsrt}
\bibliography{main}
}
\end{document}